%% file: main.tex
\documentclass[letterpaper]{article} 
\usepackage{aaai24}  
\usepackage{times}  
\usepackage{helvet}  
\usepackage{courier}  
\usepackage[hyphens]{url}  
\usepackage{graphicx} 
\usepackage{natbib}  
\usepackage{caption} 
\usepackage{algorithm}
\usepackage{algorithmic}

\usepackage{newfloat}
\usepackage{listings}
\DeclareCaptionStyle{ruled}{labelfont=normalfont,labelsep=colon,strut=off} 
\floatstyle{ruled}
\newfloat{listing}{tb}{lst}{}
\floatname{listing}{Listing}
\def\mytitle{Variance-insensitive and Target-preserving Mask Refinement for Interactive Image Segmentation}

\title{\mytitle}
\author{
    Chaowei Fang\textsuperscript{\rm 1},
    Ziyin Zhou\textsuperscript{\rm 1},
    Junye Chen\textsuperscript{\rm 2},
    Hanjing Su\textsuperscript{\rm 3},
    Qingyao Wu\textsuperscript{\rm 4},
    Guanbin Li\textsuperscript{\rm 2,5}\thanks{Corresponding author.}
}
\affiliations{
    \textsuperscript{\rm 1}School of Artificial Intelligence, Xidian University, Xi’an, China\\
    \textsuperscript{\rm 2}School of Computer Science and Engineering, Research Institute of Sun Yat-sen University in Shenzhen, Sun Yat-sen University, Guangzhou, China\\
    \textsuperscript{\rm 3}Tencent\;\;
    \textsuperscript{\rm 4}School of Software Engineering, South China University of Technology, Guangzhou, China\\
    \textsuperscript{\rm 5}GuangDong Province Key Laboratory of Information Security Technology
}

\usepackage{bibentry}

\usepackage{epsfig}
\usepackage{amsmath}
\usepackage{dsfont}
\usepackage[super]{nth}
\usepackage{multirow}
\usepackage{color}
\usepackage{amsfonts} 
\usepackage{booktabs}
\graphicspath{{./figures/}}
\begin{document}

\maketitle

\input{sections/abstract}
\input{sections/0-intro}

\input{sections/1-related}

\input{sections/2-method}
\input{sections/3-exper}

\section{Conclusion}
In this paper, we introduce a cutting-edge approach termed \textit{Variance-insensitive and Target-preserving Mask Refinement} for the point-based interactive image segmentation task. Our methodology encompasses a mask matching regularization, fortifying consistency in predictions arising from diverse initial masks. Such regularization substantially alleviates the prediction sensitivity to initial mask fluctuations. To alleviate the dilution of target information during input image downsampling, we deploy a target-aware image zooming mechanism, complementing traditional interpolation techniques. Comprehensive evaluations on datasets—GrabCut, Berkeley, SBD, and DAVIS—confirm our model's superiority over existing methods.

\section{Acknowledgements}
This work was supported in part by the National Natural Science Foundation of China (NO.~62376206, NO.~62003256, NO.~62322608), in part by the Shenzhen Science and Technology Program (NO.~JCYJ20220530141211024), and in part by the Open Project Program of the Key Laboratory of Artificial Intelligence for Perception and Understanding, Liaoning Province (AIPU, No.~20230003).

\bibliography{main}

\end{document}

%% file: sections/abstract.tex
\begin{abstract}
Point-based interactive image segmentation can ease the burden of mask annotation in applications such as semantic segmentation and image editing. However, fully extracting the target mask with limited user inputs remains challenging. We introduce a novel method, Variance-Insensitive and Target-Preserving Mask Refinement to enhance segmentation quality with fewer user inputs. Regarding the last segmentation result as the initial mask, an iterative refinement process is commonly employed to continually enhance the initial mask. Nevertheless, conventional techniques suffer from sensitivity to the variance in the initial mask. To circumvent this problem, our proposed method incorporates a mask matching algorithm for ensuring consistent inferences from different types of initial masks. We also introduce a target-aware zooming algorithm to preserve object information during downsampling, balancing efficiency and accuracy. Experiments on GrabCut, Berkeley, SBD, and DAVIS datasets demonstrate our method's state-of-the-art performance in interactive image segmentation.
\end{abstract}

%% file: sections/0-intro.tex
\begin{figure}[t]
  \begin{center}
      \includegraphics[width=0.9\linewidth]{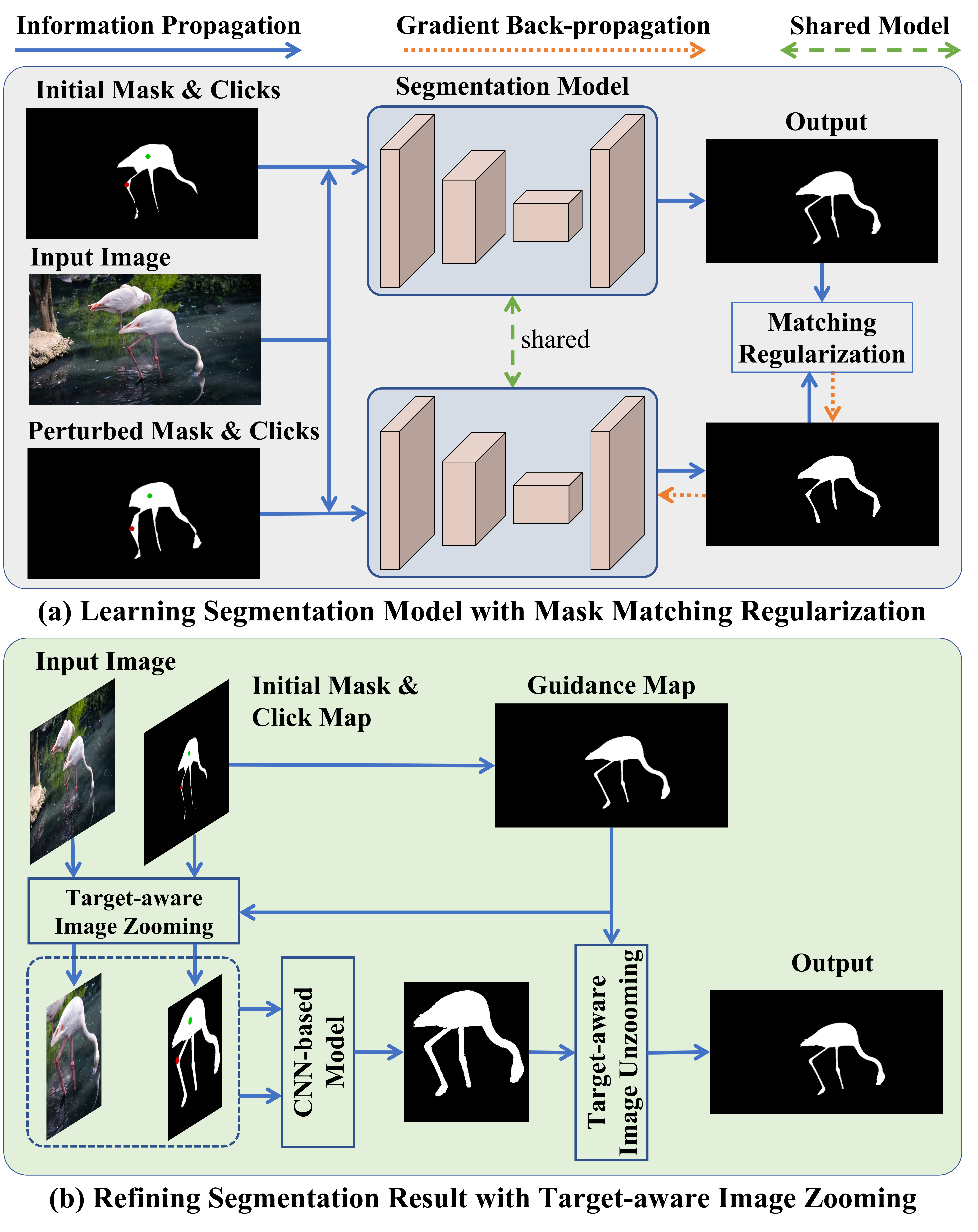}
  \end{center}
     \caption{
     We enhance the interactive image segmentation model's robustness against initial mask fluctuations using mask matching regularization (a) and introduce a target-aware zooming operation (b) for image downsampling.
     }
  \label{fig2}
\end{figure}

\section{Introduction}
Interactive Image Segmentation (IIS) serves as a prominent method for the extraction of binary masks corresponding to targeted objects, guided by user interaction cues. It holds substantial significance in diverse applications, ranging from easing the burden of data annotation in semantic segmentation to acting as essential components in image editing tasks such as image inpainting~\cite{bertalmio2000image}. 
The field of IIS recognizes a variety of interaction cues, including points~\cite{xu2016deep}, bounding boxes~\cite{yu2017loosecut}, and scribbles~\cite{lin2016scribblesup}. Within this study, our focus is drawn to the utilization of positive and negative click points (see Fig.~\ref{fig2}). 
The central challenge lies in the generation of accurate object masks with minimum clicks.

The earliest point guided IIS method built upon deep neural networks (DNN)  can be traced to \cite{xu2016deep}. Subsequent developments have introduced iterative methods to increase the flexibility of model training. For instance, \citet{mahadevan2018iteratively} unveiled an iterative training framework that autonomously samples pseudo click points relative to error maps between predicted and ground-truth (GT) segmentations, thereby enhancing the adaptability of IIS. Similarly, \citet{9897365} strategized an approach to iteratively refine the previously generated mask in the current interaction phase.
Recent contributions such as \cite{Lin_2022_CVPR} and \cite{Chen_2022_CVPR} have expanded upon this foundation by employing a coarse-to-refine framework. This methodology initially executes a preliminary coarse segmentation from a global low-resolution perspective and subsequently refines the details from a localized high-resolution viewpoint.

\begin{figure}[t]
  \begin{center}
      \includegraphics[width=1.0\linewidth]{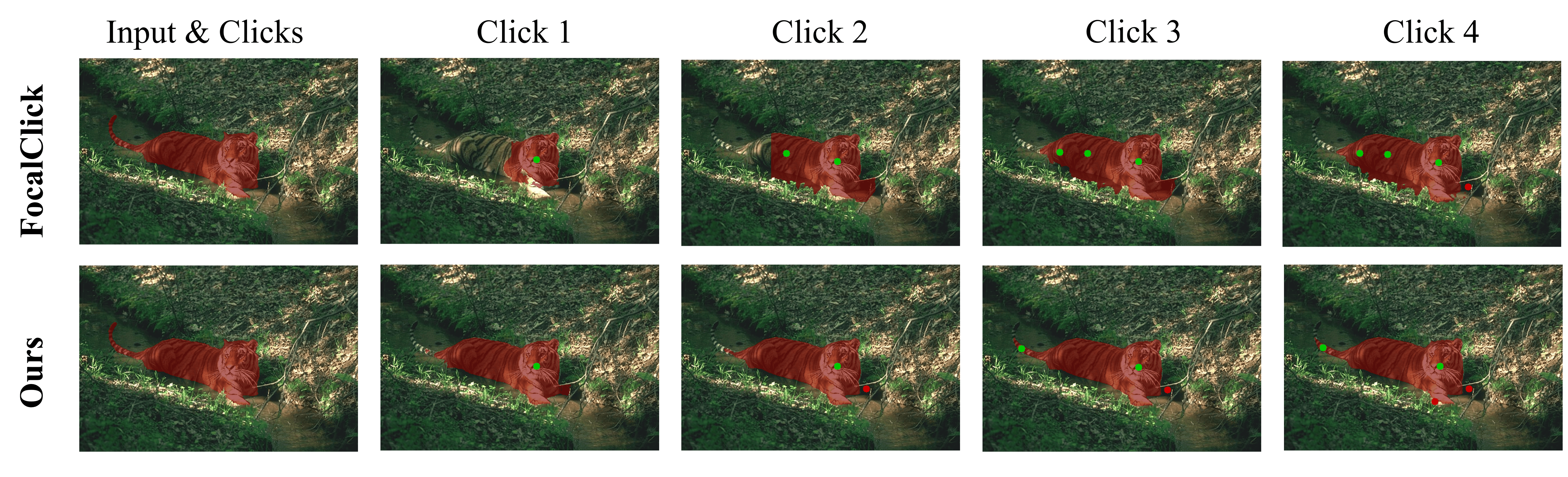}
  \end{center}
     \caption{
     Segmentation results of FocalClick and our method across varying click numbers. Green and red points indicate foreground and background clicks, respectively.
     }
  \label{fig1b}
\end{figure}

Despite recent advancements, existing methods still necessitate a moderate number of interaction points to attain satisfactory performance. A significant issue with current techniques~\cite{9897365,Lin_2022_CVPR,Chen_2022_CVPR}, is their sensitivity to the fluctuation of the initial mask. During each training step, the initial mask is either set as fully zero or derived from the model's previous prediction. Consequently, models trained in this fashion exhibit a lack of robustness when addressing the fluctuation of initial masks during inference. 
The other prevailing challenge is that existing approaches typically rely on conventional downsampling techniques, such as bilinear interpolation, to enhance inference efficiency. This process inevitably leads to information loss, complicating the discrimination of the target object. As seen in Fig.~\ref{fig1b}, a state-of-the-art method FocalClick~\cite{Chen_2022_CVPR} still requires quite a few clicks to delineate the complete object.

In response to these challenges, we introduce a novel algorithm entitled \textit{Variance-Insensitive and Target-preserving Mask Refinement}. Centered on a DNN model, our method predicts the segmentation map for an image, utilizing the click map and initial mask as supplementary inputs.
To fortify the model against initial mask fluctuation, we propose a mask matching regularization strategy. This involves generating two initial mask variants: 1) utilizing the segmentation map from the first interaction step; 2) synthesizing a mask by distorting the GT mask. Drawing inspiration from the smoothness assumption~\cite{bonaccorso2017machine}, we establish a regularization term between model predictions for the two initial mask variants (Fig.~\ref{fig2}~(a)). 
This innovation allows greater flexibility in selecting the initial mask, diversifying beyond conventional methods like RITM~\cite{ritm2022} and FocalClick~\cite{Chen_2022_CVPR}.
Furthermore, inspired by~\cite{thavamani2023learning}, we incorporate a Target-Aware Image Zooming (TAIZ) operation (Fig.~\ref{fig2}~(b)),  to mitigate information loss during image downsampling. 
Our TAIZ operation uniquely leverages the combination of the last segmentation map and click map to generate a re-sampling grid that redirects points outside the salient region to inside it.
The efficacy of our approach is evident in Fig.~\ref{fig1b}, which illustrates superior results with even fewer clicks compared to FocalClick. Comprehensive experiments conducted on four public datasets, namely GrabCut~\cite{rother2004grabcut}, Berkeley~\cite{937655}, SBD~\cite{6126343}, and DAVIS~\cite{Perazzi_CVPR_2016}, demonstrate that our method sets a new benchmark for state-of-the-art performance.

Main contributions of this paper are summarized as below.
\begin{itemize}
  \setlength\itemsep{0.5em}
  \item We develop an innovative IIS framework with the mask matching regularization. This alleviates the model's sensitivity to variances in the initial mask.
  \item We introduce a target-aware image zooming operation which can maintain the intrinsic characteristics of the target object during the input image downsampling process.
  \item We conduct extensive experiments on GrabCut, Berkeley, SBD, and DAVIS datasets. The results affirm that our method significantly surpasses existing methods.
\end{itemize}

%% file: sections/1-related.tex
\section{Related Work}
\subsection{Interactive Image Segmentation}
Interactive image segmentation (IIS) has been a longstanding challenge in computer vision. 
The advent of deep learning in semantic segmentation led to its application in IIS by \citet{xu2016deep}, establishing a mainstream approach.
Early deep learning-based IIS methods~\cite{8237559, xu2016deep, xu2017deep} overlooked the information contained in previously generated masks. \citet{mahadevan2018iteratively} recognized the importance of previous segmentation results as additional inputs, a concept subsequently adopted by many researchers \cite{Lin_2020_CVPR, Chen_2022_CVPR, wei2023focused, zhou2023interactive}.

\citet{hao2021edgeflow} sought to fully utilize generated masks by implementing multi-stage feature fusion, while others performed a coarse segmentation and then incorporated additional modules to refine the coarse segmentation output~\cite{Chen_2021_ICCV, Hao_2021_ICCV}. However, such approaches may substantially increase the inference time.
To accelerate the mask refinement process, \citet{wei2023focused} attempted to accelerate local refinement through similarity-driven updates, and~\citet{Chen_2022_CVPR} exploited local refinement by focusing on specific regions. 

However, existing methods suffer from the loss of critical visual information during the downsampling process, which is typically employed to ensure efficiency in inference. This loss is particularly detrimental to identifying intricate aspects of the target object such as boundaries and small-scale components.
To address this challenge, we introduce a new target-aware image zooming (TAIZ) algorithm. 
Unlike traditional downsampling methods, TAIZ accentuates the content of the target object, thus offering a more nuanced understanding of the target object.

\subsection{Consistency Regularization}
In line with the smoothness hypothesis~\cite{bonaccorso2017machine}, a model should exhibit robustness against variations in the input, meaning that the introduction of noise to input samples should not significantly affect the model's inference. This principle has inspired many semi-supervised learning techniques e.g.~\cite{xie2020unsupervised,sajjadi2016regularization}. 
It also has important implications for IIS methods~\cite{9897365} based on iterative mask refinement. Such methods are sensitive to variations in the initial mask.
To address this issue, we propose a mask matching regularization strategy, which enhances model robustness by enforcing the model to generate consistent predictions across different variations of the initial mask. 

%% file: sections/2-method.tex
\begin{figure*}[t]
  \begin{center}
      \includegraphics[width=0.92\linewidth]{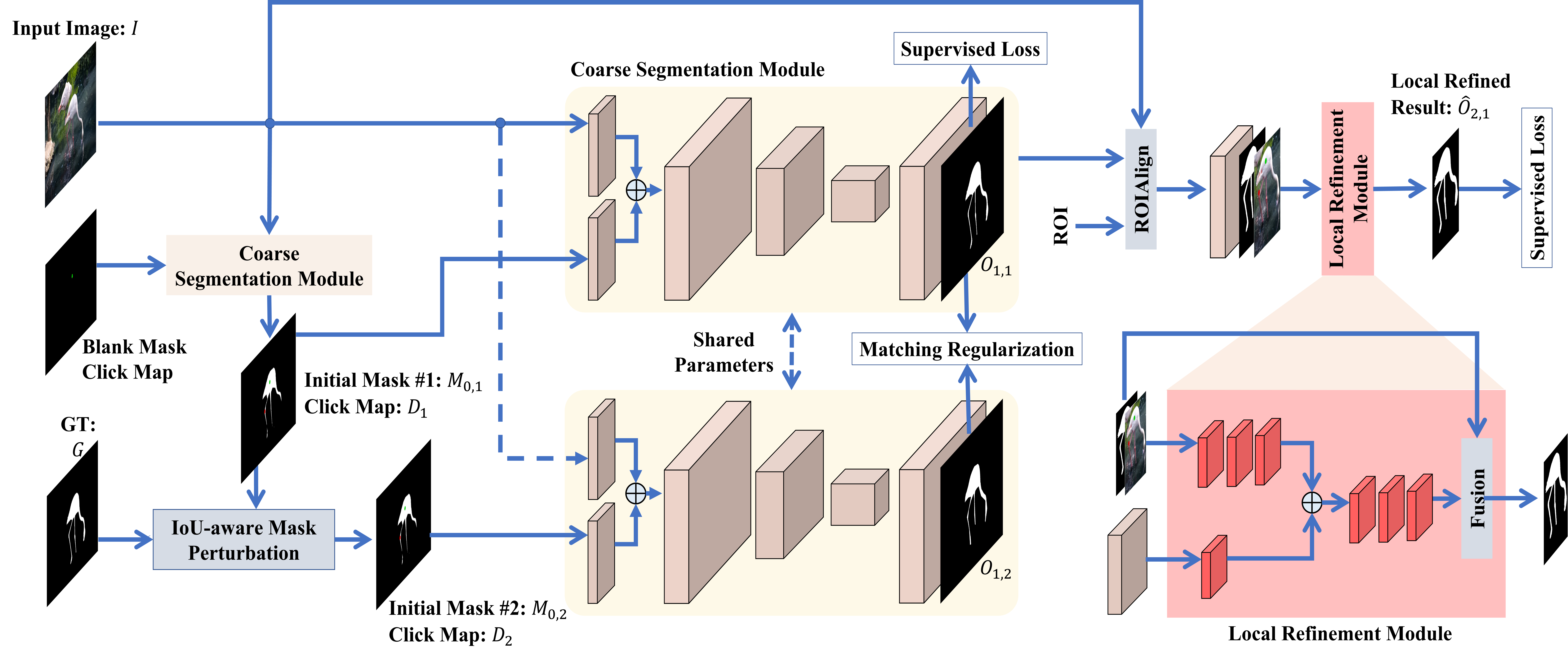}
  \end{center}
     \caption{Workflow of our method during training. The network architecture is composed of a coarse segmentation module and a refinement module. Two types of initial masks are adopted to construct the mask matching regularization. 
      }
  \label{fig:long}
  \label{pipeline}
\end{figure*}

\section{Methodology}
This work addresses the problem of point-based interactive image segmentation. Given an image $I \in \mathbb{R}^{H \times W \times 3}$, we train a DNN model to extract the target object through $T$ interaction steps. In addition to the image and cumulative click points, the inputs include the segmentation map obtained from the previous interaction step as the initial mask. 

\subsection{Framework Overview}
As shown in Fig.~\ref{pipeline}, we follow~\cite{Chen_2022_CVPR} to construct the network architecture which is consisting of a coarse segmentation module and a local refinement module. 
The coarse segmentation module generates a segmentation map from the input image $I$, guided by an initial mask and click map. 
Let the initial mask be denoted by $M_0 \in \mathbb R^{H \times W}$, and the click map be represented as a two-dimensional disc map $D \in \mathbb R^{H \times W \times 2}$, indicating the positions of positive and negative clicks. 
The segmentation logit of the coarse segmentation module is defined as $O_1 \in \mathbb R^{H \times W}$, where $O_1 = \mathcal F_{coarse}(I, D, M_0)$ and $\mathcal F_{coarse}(\cdot)$ denotes the inference function of the coarse segmentation module. The coarse segmentation map $M_1$ is then obtained by thresholding $O_1$.
The local refinement module is targeted at improving the segmentation in a specific local region determined by the maximum connected region in the difference map between $M_1$ and $M_0$. Concretely, it extracts patches from $I$, $O_1$, $D$, and the penultimate feature map of the coarse segmentation module, corresponding to the local region. Then, it regards theses patches as inputs, generating refined local segmentation logit $\hat O_1$ which is subsequently used for updating the coarse segmentation result.

To fortify the robustness against fluctuations in the initial mask, we introduce a regularization approach that ensures consistency between coarse segmentation results derived from different forms of initial masks. Additionally, we design a target-aware image zooming algorithm to retain the target content while downsampling the input image.

\subsection{Learning with Mask Matching Regularization}

In point-based IIS, a common approach is the iterative refinement of the current segmentation map with newly incorporated clicks~\cite{9897365,Lin_2022_CVPR}. While this iterative pipeline efficiently leverages previously generated segmentation results, it often lacks robustness against changes in the initial mask, as it relies on either a blank mask or the segmentation map of the last interaction step during training.

To mitigate this limitation, we introduce a novel regularization approach called \textit{Mask Matching Regularization} (MaskMatch). 
Specifically, each training sample comprises four elements: the input image $I$, initial mask $M_0$, click map $D$, and ground-truth segmentation map $G$.
Following~\cite{Chen_2022_CVPR}, new positive (negative) clicks are synthesized from false negative (false positive) pixels in $M_0$.
Next, two temporary masks $M_{0,1}^\prime$ and $M_{0,2}^\prime$ are generated according to two distinct strategies:
1) We input $I$, a blank mask, and a synthesized click point into the coarse segmentation module, yielding $M_{0,1}^\prime$.
2) $M_{0,2}^\prime$ is generated by perturbing $G$ with boundary adjustment and region interference operations as in~\cite{cheng2020cascadepsp} continuously, until it reaches the IoU value of $M_{0,1}^\prime$.

With $M_{0,1}^\prime$ and $M_{0,2}^\prime$, the interactive segmentation process restarts for $K$ additional steps (where $K$ is randomly chosen from $\{0,1,2,3\}$),  producing two segmentation masks $M_{0,1}$ and $M_{0,2}$, respectively. 
Two click maps $D_1$ and $D_2$ can be acquired according to $M_{0,1}$ and $M_{0,2}$, respectively.
These elements are then fed into the coarse segmentation module to generate two segmentation logits $O_{1,1} =\mathcal F_{coarse}(I, D_1, M_{0,1})$ and $O_{1,2} =\mathcal F_{coarse}(I, D_2, M_{0,2})$.
A matching regularization term between $O_{1,1}$ and $O_{1,2}$ is then established:
\begin{equation} \label{eq:loss-cr}
L_{mr} = \Gamma(\mathds{1}(\sigma(O_{1,1})>0.9) \circ \ell_{bce}(\sigma(O_{1,2}),\sigma(O_{1,1}) )),
\end{equation}
where $\circ$ denotes the element-wise product; 
$\mathds{1}(\cdot)$ is the indicator function; $\sigma(\cdot)$ is the Sigmoid function; $\ell_{bce}(\cdot)$ represents the binary cross entropy function; $\Gamma(\cdot)$ is the element-wise average function.
The optimization process is stabilized by back-propagating the gradient through $O_{1,2}$ but not $O_{1,1}$, and selecting only high-confidence pixels in $O_{1,1}$ for the computation of the regularization term.

The conventional supervised learning objective constrains predictions on $(I, D_1, M_{0,1})$: 
\begin{equation}
  \begin{aligned}\label{eq:loss-sup}
  L_{sup} = & \Gamma(\ell_{nf}(\sigma(O_{1,1}), G))\\
    &+\Gamma(\ell_{nf}(\sigma(\hat O_{1,1}), \hat G) + \ell_{nf}(\hat E_{1,1}, \hat G_e) ),
  \end{aligned}
\end{equation}
where $\hat O_{1,1}$ is the segmentation logit of the refinement module; $\hat G$ is the GT mask of the local view; $\hat E_{1,1}$ is the edge map predicted by an auxiliary branch in the refinement module following~\cite{Chen_2022_CVPR}, and $\hat G_e$ represents the GT of the edge map; $\ell_{nf}(\cdot)$ denotes the normalized focal loss function proposed in~\cite{sofiiuk2019adaptis}.

The overall loss function is formed by combining these two terms in Eq.~(\ref{eq:loss-cr}) and~(\ref{eq:loss-sup}):
\begin{equation}
L = L_{sup} + \mathds{1}(\text{IoU}(M_{0,1}, G)>\alpha)\times L_{mr}, 
\end{equation}
where $\alpha$ ($=0.8$) is a constant;
$\text{IoU}(M_{0,1}, G)$ calculates the intersection-over-union between $M_{0,1}$ and $G$. Here, if the quality of $M_{0,1}$ is not high, the mask matching regularization would be ignored, since $O_{1,1}$ may not be able to provide accurate supervision under such circumstance.

\begin{figure}[t]
  \begin{center}
      \includegraphics[width=1\linewidth, trim={20 20 20 20},clip]{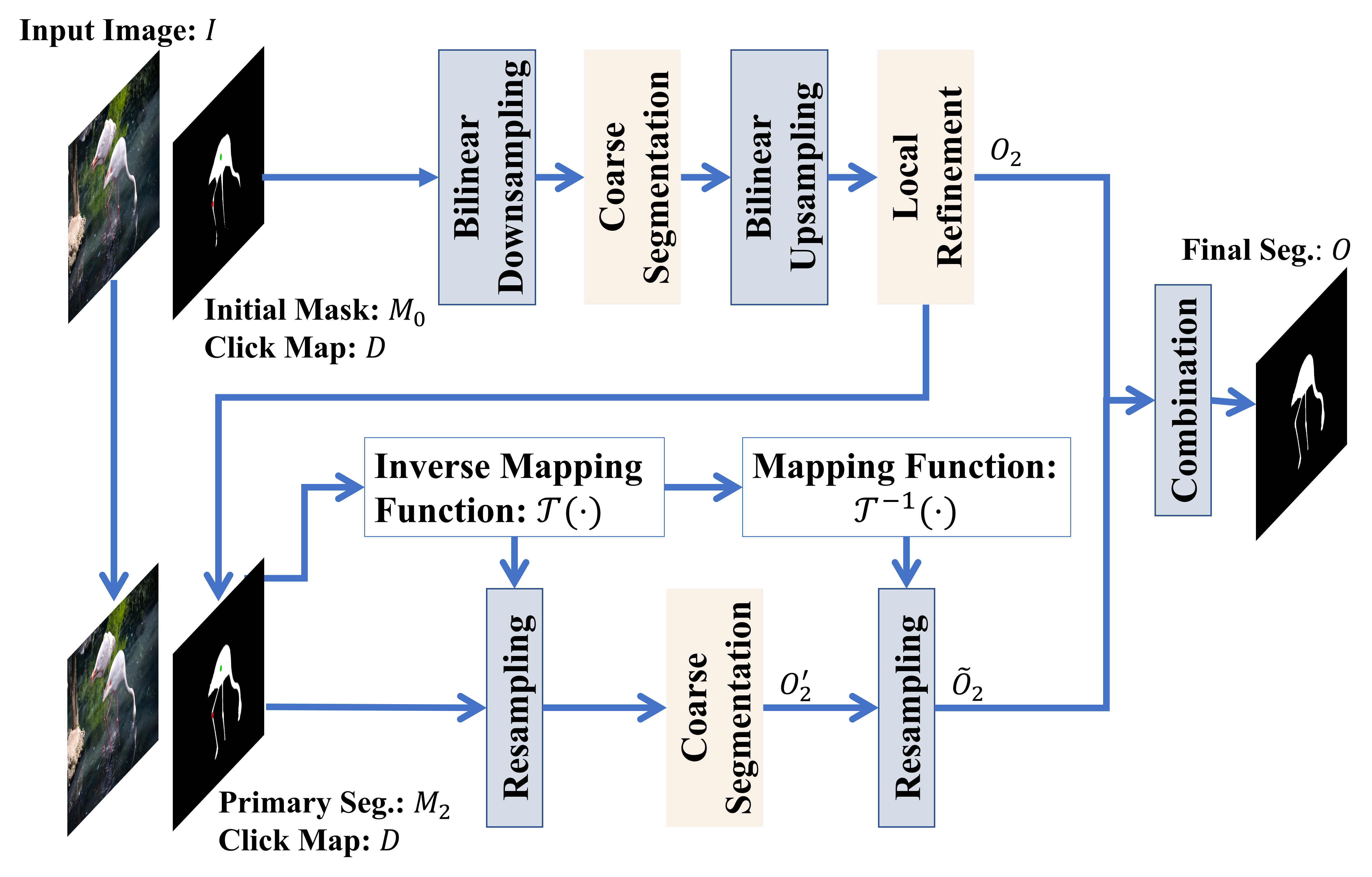}
  \end{center}
     \caption{Inference process of our method. 
      }
  \label{fig:infer}
\end{figure}

\subsection{Target-Aware Image Zooming } 
\label{sec:taiz}

Traditional methods often downsample images using standard interpolation algorithms to save computation but compromise essential visual details. Drawing inspiration from~\cite{thavamani2023learning}, we introduce \textit{Target-Aware Image Zooming} (TAIZ), ensuring the preservation of crucial object information while reducing image resolution.

The TAIZ operation counteracts the loss of visual details by employing denser pixel sampling in areas of interest.
We initiate by defining a guidance map \(S \in \mathbb{R}^{H \times W}\) which specifies pixel-wise locations for the target object. An inverse mapping function \(\mathcal{T}: [0,1]^2 \rightarrow [0,1]^2\) is constructed to map a point \((x,y)\) in the target image to \((\mathcal{T}_x(x), \mathcal{T}_y(y))\) in the source image. \(\mathcal{T}_x(x)\) and \(\mathcal{T}_y(y)\) determine the horizontal and vertical coordinates, respectively.
Marginalizing \(S\) horizontally and vertically yields vectors \(S_y \in \mathbb{R}^H\) and \(S_x \in \mathbb{R}^W\) respectively, which are represented as \(S_y = S \cdot \mathbf{1}^{W \times 1}\) and \(S_x = (\mathbf{1}^{1 \times H} \cdot S)^{\text{T}}\). Here, $S_y$ and $S_x$ reflect the importance levels of rows and columns, respectively.
Then, \(\mathcal{T}_x(x)\) and \(\mathcal{T}_y(y)\) are calculated as below:
\begin{align}
  \mathcal{T}_x(x) & = \frac{\int_{x'} x' S_x(x') \mathcal{K}(x, x') \, dx'}{\int_{x'} S_x(x') \mathcal{K}(x, x') \, dx'}, \\
  \mathcal{T}_y(y) & = \frac{\int_{y'} y' S_y(y') \mathcal{K}(y, y') \, dy'}{\int_{y'} S_y(y') \mathcal{K}(y, y') \, dy'},
\end{align}
where \(\mathcal{K}(x, x') = e^{-\frac{(x-x')^2}{2\sigma^2}}\) (with \(\sigma\) as the standard deviation) is the Gaussian kernel function. 
Owing to the weight modulation from $S_x$ and $S_y$, this inverse mapping function concentrates insignificant points towards salient ones in the guidance map \(S \). 
Hence, it can be used to downsample images without substantial information loss for the salient regions implied by \(S \) as visualized in Fig.~\ref{fig2} (b).

During training, the above TAIZ operation is used to distort half of images by regarding the GT mask as the guidance map, optimizing segmentation performance on TAIZ-processed images.
During testing, the input image $I$, click map $D$, and initial mask $M_0$ are inputted into the interaction segmentation pipeline, producing a segmentation logit $O_2$. Bilinear interpolation is used to downsample inputs for efficiency. 
Applying a threshold of 0 to $O_2$  yields mask $M_2$.
The union of $M_2$ and $D$ serves as the guidance map for creating the inverse mapping function $\mathcal{T}(\cdot)$.
Using this function, we obtain the low-resolution versions of $I$, $M_2$ and $D$ as \(I'\), \(M'_2\), and \(D'\), respectively.
These are then fed into the coarse segmentation module, producing \(O'_2\).
This is resampled to the original space using \(\mathcal{T}^{-1}\), creating \(\tilde{O}_2\).
The final segmentation logit, \(O\), is generated by combining $O_2$ and \(\tilde{O}_2\) with the following formulation: \(O = (1-\lambda_t) O_2 + \lambda_t \tilde{O}_2\), where $t$ is the interaction round. Considering the quality of guidance map is not high in early interaction rounds, we set $\lambda_t$ to 0 if $t<T/2$; otherwise, \( \lambda_t = \max(T/2,t)/T\).
The inference process is depicted in Fig~\ref{fig:infer}.

%% file: sections/3-exper.tex
\begin{table*}[h]
  \centering
  \fontsize{9}{10}\selectfont

  \caption{Performance of interactive image segmenation methods evaluated with NoC metrics on GrabCut, Berkeley, SBD, and DAVIS. Lower metric values indate better performance, and the best results are indicated by bold digits. }
  \setlength{\tabcolsep}{2mm}{
    \begin{tabular}{l|c|c|c|c|c|c|c|c}
    \toprule
    {\multirow{2}[0]{*}{\textbf{Method}}}  & \multicolumn{2}{c|}{GrabCut} & \multicolumn{2}{c|}{Berkeley} & \multicolumn{2}{c|}{SBD} & \multicolumn{2}{c}{DAVIS} \\
    \cline{2-9}    
    {}  & \textcolor[rgb]{ 0,  0,  0}{NoC@85} & \textcolor[rgb]{ 0,  0,  0}{NoC@90} & \textcolor[rgb]{ 0,  0,  0}{NoC@85} & \textcolor[rgb]{ 0,  0,  0}{NoC@90} & \textcolor[rgb]{ 0,  0,  0}{NoC@85} & \textcolor[rgb]{ 0,  0,  0}{NoC@90} & \textcolor[rgb]{ 0,  0,  0}{NoC@85} & \textcolor[rgb]{ 0,  0,  0}{NoC@90} \\
    \midrule
    RIS-Net
    &    -  & $5.00$ &     - & $6.03$ &    -  &    -  &    -  &   - \\
    \hline
    LD-vgg19
    & $3.20$ & $4.79$ &     - &      - &    -  &    -  & $5.95$ & $9.57$ \\
    \hline
    CAG-fcn8s
    &    -  & $3.58$ &     - & $5.60$ &    -  &    -  &    -  &   - \\
    \hline
    BRS-densenet
    & $2.60$ & $3.60$ &     - & $5.08$ & $6.59$ & $9.78$ & $5.58$ & $8.24$ \\
    \hline
    FCA-resnet101
    &     -    & $2.24$ &     - & $4.23$ &    -  &    -  &     - & $7.90$ \\
    FCA-res2net
    &     - & $2.08$ &     - & $3.92$ &     - &     - &    -  & $7.57$ \\
    \hline
    f-BRS-resnet101
    & $2.30$ & $2.78$ &     - & $4.57$ & $4.81$ & $7.73$ & $5.04$ & $7.81$ \\
    \hline
    CDNet-resnet101
    & $2.42$ & $2.76$ &     - & $3.65$ & $4.73$ & $7.66$ & $5.33$ & $6.97$ \\
    \hline
    RITM-hrnet18s
    & $1.54$ & $1.68$ &     - & $2.6$ & $4.04$ & $6.48$ & $4.7$ & $5.98$ \\
    RITM-hrnet18
    & $1.42$ & $1.54$ &     - & $2.26$ & $3.80$ & $6.06$ & $4.36$ & $5.74$ \\
    RITM-hrnet32
    & $1.46$ & $1.56$ &     - & $2.10$ & $3.59$ & $5.71$ & $4.11$ & $5.34$ \\
    \hline
    EdgeFlow-hrnet18
    & $1.60$ & $1.72$ &     - & $2.40$ &     - &     - & $4.54$ & $5.77$ \\
    \hline
    FICI-hrnet18s
    & $1.50$ & $1.56$ &  - & $2.05$ & $3.88$ & $6.24$ & $3.7$ & $5.16$ \\
    FICI-hrnet18
    & \textbf{$1.38$} & $1.46$ &     -  & $1.96$ & $3.63$ & $5.83$ & $3.97$ & $5.16$ \\
    \hline
    FocalClick-hrnet18s
    & $1.48$ & $1.62$ & $1.60$ & $2.23$ & $4.43$ & $6.79$ & $3.90$ & $5.23$ \\
    FocalClick-segformerB0
    & $1.40$ & $1.66$ & $1.59$ & $2.27$ & $4.56$ & $6.86$ & $5.04$ & $5.49$ \\
    FocalClick-segformerB3
    & $1.44$ & $1.50$ & $1.55$ & $1.92$ & $\bf{3.53}$ & $5.59$ & $3.61$ & $4.90$ \\

    \hline

    Ours-segformerB0  
    & $1.42$ & $1.54$ & $1.64$ & $2.18$ & $4.43$ & $6.75$ & $3.81$ & $5.39$ \\
    Ours-segformerB3  
    &  $\textbf{1.38}$ & $\bf{1.42}$ & $\bf{1.44}$ & $\bf{1.72}$ & $3.55$ & $\bf{5.53}$ & $\bf{3.26}$ & $\bf{4.82}$ \\
    \bottomrule
    \end{tabular}
    }
  \label{compareWithSOTA}%
\end{table*}%

\begin{table*}[h]
  \centering
  \fontsize{9}{10}\selectfont
  \caption{Performance of IIS methods evaluated with segmentation quality and efficiency metrics on Berkeley and DAVIS datasets. `$\downarrow$' (`$\uparrow$') means lower (higher) metric values indicate better performance.
  }
  \setlength{\tabcolsep}{2mm}{
    \begin{tabular}{l|c|c|c|c|c|c|c|c}
    \toprule
    {\multirow{2}[0]{*}{\textbf{Method}}} & \multicolumn{4}{c|}{ Berkeley } & \multicolumn{4}{c}{DAVIS} \\
    \cline{2-9}
    {} & \textcolor[rgb]{ 0,  0,  0}{NoF@90 $\downarrow$} & \textcolor[rgb]{ 0,  0,  0}{IoU@5 $\uparrow$} & \textcolor[rgb]{ 0,  0,  0}{BIoU@5 $\uparrow$ } & SPC $\downarrow$   & \textcolor[rgb]{ 0,  0,  0}{NoF@90 $\downarrow$} & \textcolor[rgb]{ 0,  0,  0}{IoU@5 $\uparrow$} & \textcolor[rgb]{ 0,  0,  0}{BIoU@5 $\uparrow$} & SPC $\downarrow$ \\
    \midrule
    {f-BRS-B-resnet101} & $6$ & $0.875$ & $0.73$ & $0.072$ & $77$ & $0.826$ & $0.717$ & $0.102$ \\
    \hline
    {FCA-Net-resnet101} & $7$ & $0.923$ & $0.793$ & $0.059$ & $74$ & $0.867$ & $0.771$ & $0.075$ \\
    \hline
    {CDNet-resnet101} & $4$ & $0.921$ & $0.803$ & $0.079$ & $60$ & $0.876$ & $0.783$ & $0.108$ \\
    \hline
    {FICI-hrnet18s} & $0$ & $0.958$ & $0.883$ & $0.044$ & $51$ & $0.907$ & $0.830$ & $0.069$ \\
    \hline
    {FocalClick-segformerB0} & 2 & 0.957 & 0.798 & $\textbf{0.017}$ & 54 & 0.903 & 0.724 & $\textbf{0.024}$ \\
    {FocalClick-segformerB3} & $\textbf{0}$ & $0.962$ & $0.896$ & 0.037 & $\bf{50}$ & $0.912$ & $0.841$ & 0.048 \\
    \hline
    {Ours-segformerB0} & $\textbf{0}$ & 0.961 & 0.810 & 0.027 & 51 & 0.912 & 0.748 & 0.040 \\
    {Ours-segformerB3} & $\textbf{0}$ & $\textbf{0.963}$ & $\textbf{0.897}$ & $0.054$ & $\bf{50}$ & $\textbf{0.916}$ & $\textbf{0.848}$ & $0.067$ \\
    \bottomrule
    \end{tabular}%
  }
  \label{tradeoff}%
\end{table*}%

\section{Experiments}
\subsection{Datasets and Evaluation Metrics}

{\bf Datasets.} 
The training images are collected from COCO~\cite{10.1007/978-3-319-10602-1_48} and LVIS~\cite{Gupta_2019_CVPR} datasets, containing $1.04\times 10^5$ images and 1.6 million instance-level masks. Four publicly available datasets are used for evaluating IIS methods:

\begin{itemize} \setlength\itemsep{0.2em} 
\item \textbf{GrabCut\cite{rother2004grabcut}} contains 50 images with single object masks. 
\item \textbf{Berkeley \cite{937655}} contains 96 images with 100 object masks. 
\item \textbf{SBD \cite{6126343}} is comprised of 8,498 training images with 20,172 polygonal masks, and 2,857 validating images with 6,671 instance-level masks. Only the validating images are used for evaluation. 
\item \textbf{DAVIS \cite{Perazzi_CVPR_2016}} contains 345 frames randomly sampled from 50 videos. Each frame is provided with high-quality masks. 
\end{itemize}

\begin{figure*}[h]
  \begin{center}
      \includegraphics[width=0.92\linewidth]{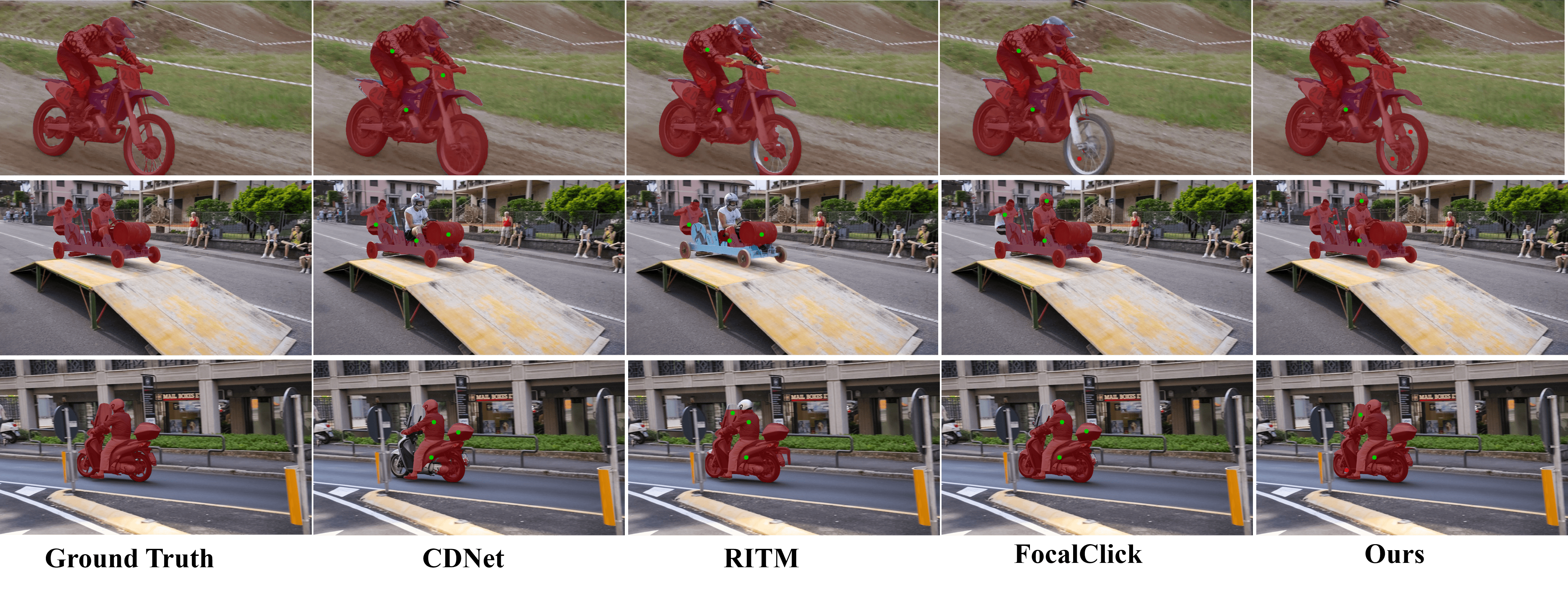}
  \end{center}
     \caption{Qualitative comparisons of CDNet, RITM, FocalClick, and our method. 
     }
  \label{figquan}
\end{figure*}

\begin{figure}[t]
  \begin{center}
      \includegraphics[width=0.92\linewidth]{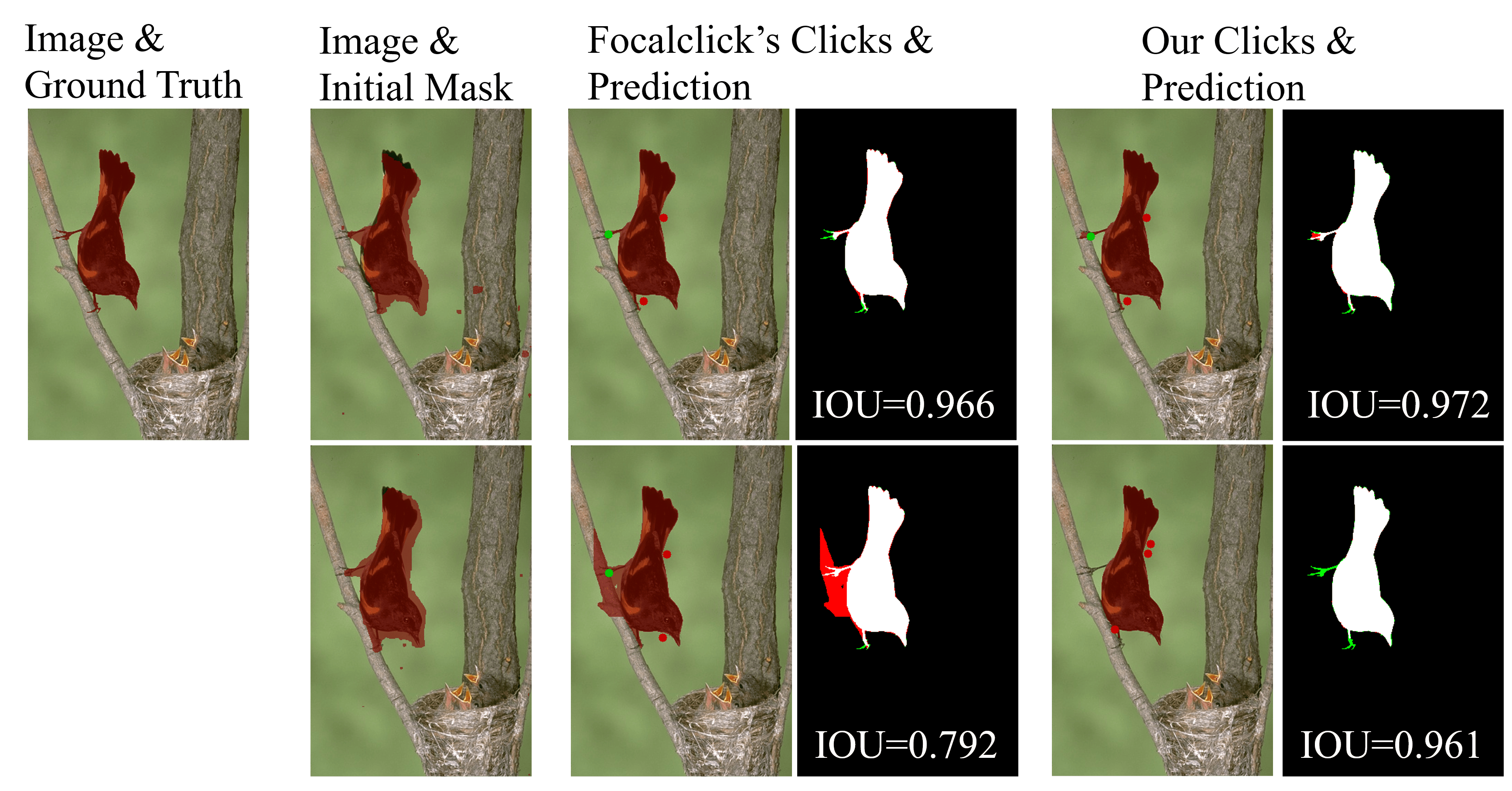}
  \end{center}
     \caption{Comparison of the results obtained from the FocalClick and our method with different initial masks. In the 4th and 6th columns, white, red, and green indicate true positives, false positives, and false negatives, respectively. 
     }
  \label{figmaskmatch}
\end{figure}

\begin{figure}[t]
  \begin{center}
      \includegraphics[width=0.92\linewidth]{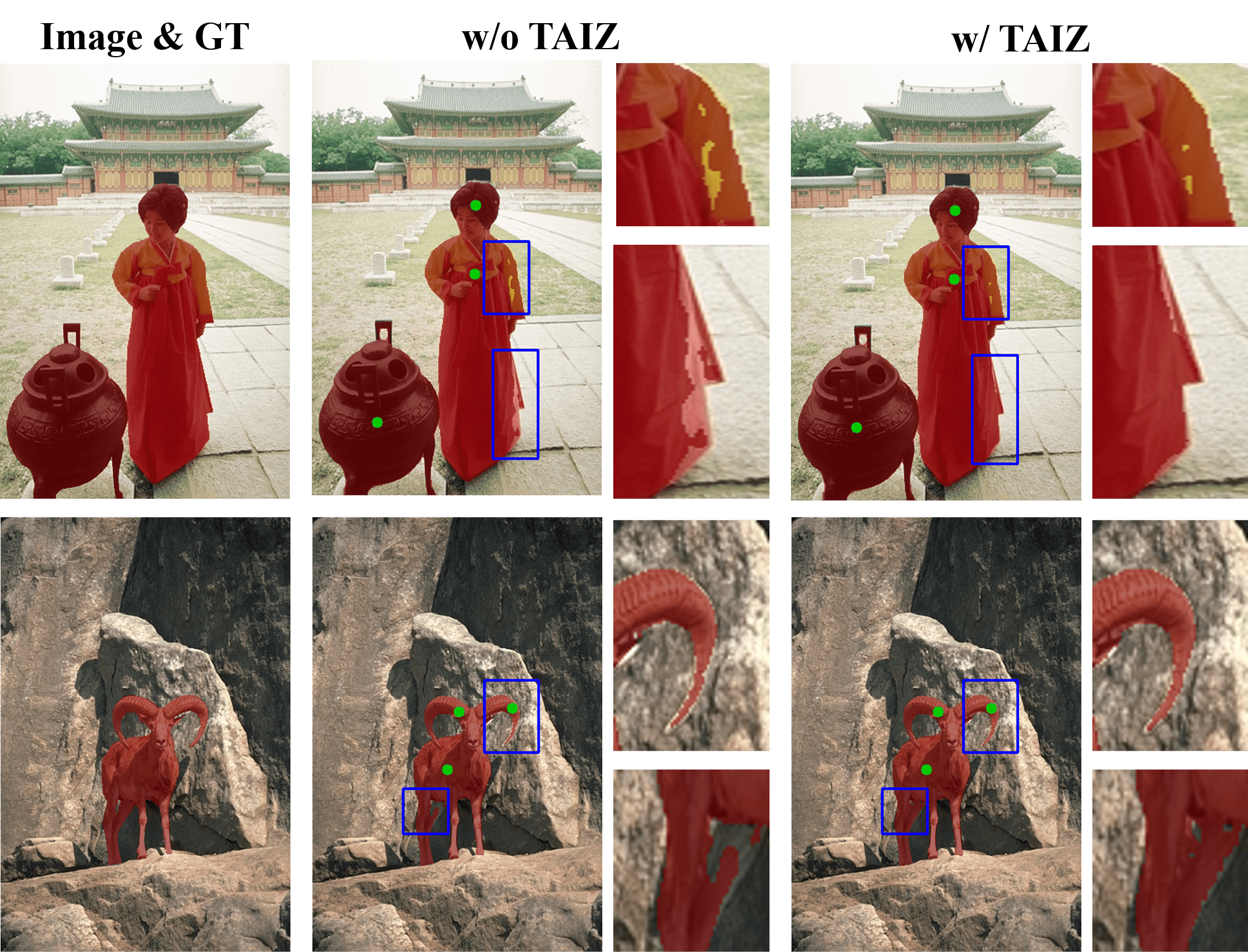}
  \end{center}
     \caption{Qualitative comparison between method variations using or not using TAIZ algorithm.}
  \label{figmaskzoom}
\end{figure}

\noindent{\bf Experimental Setting.}
We choose segformerB0 or segformerB3~\cite{xie2021segformer} as the backbone of the segmentation model. 
During the training phase, in accordance with the approach described in~\cite{Chen_2022_CVPR}, 30,000 images are randomly selected as the training dataset for each epoch.
Initially, the images are downsampled using either bilinear interpolation or TAIZ. Data augmentation is subsequently applied, encompassing random flipping, resizing with a scale factor constrained within the interval $[0.75, 1.40]$, and randomized adjustments to brightness, contrast, and RGB coloration. $\sigma$ is set to 11.
The network parameters are optimized through the Adam algorithm, parameterized with $\beta_1=0.9$ and $\beta_2=0.999$. The model undergoes a training process of 230 epochs, with an initial learning rate of $5\times 10^{-4}$. This learning rate is subsequently attenuated by a factor of $0.1$ at the 190th and 220th epochs.
Training is executed with a batch size of 24, using PyTorch as the implementation framework. All computational experiments are performed on a system equipped with two NVIDIA GeForce RTX 3090 GPUs, and the training duration for the proposed method is approximately 48 hours.

\noindent {\bf Evaluation Metrics.} 
In assessing IIS methods, we adhere to the evaluation mechanism delineated in~\cite{9897365, Chen_2022_CVPR}. The maximum click number $T$ is set to 20. The performance is quantitatively measured through five specific metrics:
\begin{enumerate}
\item 
NoC@IoU: Reflects the average number of clicks necessary to attain the specified IoU threshold.
\item NoF@IoU: Quantifies the number of instances where the model fails to reach the prescribed IoU threshold within the maximum allowable number of clicks.
\item IoU@$N$: Denotes the mean IoU achieved for testing images after $N$ clicks.
\item BIoU@$N$: Signifies the mean boundary IoU of the testing images after $N$ clicks.
\item SPC: Represents the mean computational time required for inference following each click.
\end{enumerate}

These metrics collectively provide a comprehensive and robust evaluation of the model's efficacy and efficiency.

\subsection{Comparison with Other Methods}

In our comparative analysis presented in Table~\ref{compareWithSOTA}, our method is benchmarked against existing techniques, including RIS-Net~\cite{8237559}, LD~\cite{Li_2018_CVPR}, CAG~\cite{Majumder_2019_CVPR}, BRS~\cite{Jang_2019_CVPR}, FCA~\cite{Lin_2020_CVPR}, f-BRS~\cite{Sofiiuk_2020_CVPR}, CDNet~\cite{Chen_2021_ICCV}, RITM~\cite{9897365}, EdgeFlow~\cite{Hao_2021_ICCV}, FICI~\cite{wei2023focused}, and FocalClick, using the NoC metrics. Owing to the incorporation of the mask matching regularization and TAIZ operation, our proposed algorithm consistently necessitates fewer clicks to achieve IoU thresholds of 85\% and 90\%. This establishes its superiority over contemporary state-of-the-art methods. 
Notably, the Berkeley and DAVIS datasets manifest significant enhancements when processed by our method. For instance, based on the segformerB3, we achieve reductions of 0.11 and 0.35 in NoC@85 for Berkeley and DAVIS, respectively,  relative to the prevailing state-of-the-art methods. Correspondingly, NoC@90 witnesses declines of 0.2 and 0.08 on these datasets.

Complementary metrics, encompassing NoF@90, IoU@5, BIoU@5, and SPC on the Berkeley and DAVIS datasets, are cataloged in Table~\ref{tradeoff}. Our approach consistently achieves superior values in segmentation quality metrics like IoU@5 and BIoU@5, without imposing significant time overheads. Specifically, compared to the second-ranking FocalClick, our approach enhances the BIoU@5 metric by 0.007 on DAVIS when using segformerB3 as the backbone. The SPC for our approach based on segformerB3 is 0.054s for Berkeley and 0.067s for DAVIS, indicating a computation time that is congruent with real-time applications. 
Compared to FocalClick, our method maintains the same model complexity and brings minimal additional computational or memory overhead during inference.
Fig.~\ref{figquan} offers a visual depiction of the results. Relative to competitive techniques, our method yields more refined segmentations with enhanced boundary fidelity.


\noindent \textbf{Robustness against Fluctuation in Initial Mask.}
To assess the resilience of our approach to fluctuation in the initial mask, we introduce perturbations to the GT mask of each test image ten times, guided by a threshold of 0.8 IoU. 
The capability of FocalClick and our method to rectify errors on these perturbed masks is detailed in Table~\ref{initmask}. A visual representation comparing the outcomes of FocalClick and our approach for two initial masks is depicted in Fig.~\ref{figmaskmatch}. Both quantitative and qualitative evaluations corroborate that our technique consistently delivers more stable and refined segmentation results irrespective of mask initialization.

\begin{table}[t]
  \centering
  \fontsize{9}{10}\selectfont
  \caption{Average performance on correcting initial masks.}
    \begin{tabular}{l|c|c|c|c}
    \toprule
    {\multirow{2}[0]{*}{\textbf{Method}}} & \multicolumn{2}{c|}{Berkeley} & \multicolumn{2}{c}{DAVIS} \\
    \cline{2-5}    
    {} & \textcolor[rgb]{ 0,  0,  0}{NoC@85} & \textcolor[rgb]{ 0,  0,  0}{NoC@90} & \textcolor[rgb]{ 0,  0,  0}{NoC@85} & \textcolor[rgb]{ 0,  0,  0}{NoC@90} \\
    \midrule
    {FocalClick} & $1.98$ & $2.50$ & $3.97$ & $5.46$ \\
    \hline
    {Ours} & $1.86$ & $2.38$ & $3.82$ & $5.31$ \\
    \bottomrule
    \end{tabular}
  \label{initmask}%
\end{table}%

\begin{table}[t]
  \centering
  \fontsize{9}{10}\selectfont
  \caption{Ablation study for core components of our method.}
  \setlength{\tabcolsep}{0.5mm}{
    \begin{tabular}{l|c|c|c|c}
      \toprule
      {\multirow{2}[0]{*}{\textbf{Method}}} & \multicolumn{2}{c|}{Berkeley} & \multicolumn{2}{c}{DAVIS} \\
      \cline{2-5}
     {} &{\textcolor[rgb]{ 0,  0,  0}{NoC@85}} & {\textcolor[rgb]{ 0,  0,  0}{NoC@90}} & {\textcolor[rgb]{ 0,  0,  0}{NoC@85}} & {\textcolor[rgb]{ 0,  0,  0}{NoC@90}} \\
      \midrule
       {baseline} & $1.55$  & $1.92$  & $3.61$  & $4.90$ \\
      {TAIZ}                             & $1.53$  & $1.90$  & $3.54$  & $4.90$ \\
      {MaskMatch}                              & $\textbf{1.43}$  & $1.84$  & $3.48$  & $4.89$ \\
      {Replace $L_{mr}$ with $L_{sup}$}        & $1.51$  & $1.78$  & $3.49$  & $4.88$ \\
       {Ours}                                  & $1.44$  & $\textbf{1.72}$  & $\textbf{3.26}$  & $\textbf{4.82}$ \\
      \bottomrule
    \end{tabular}%
  \label{module}%
}
\end{table}%

\begin{table}[t]
  \centering
  \fontsize{9}{10}\selectfont
  \caption{Performance of our method using different thresholds to activate the mask matching regularization.}
  \setlength{\tabcolsep}{1mm}{
  \begin{tabular}{l|c|c|c|c}
    \toprule
    {\multirow{2}[0]{*}{\textbf{Threshold}}} & \multicolumn{2}{c|}{Berkeley} & \multicolumn{2}{c}{DAVIS} \\
    \cline{2-5}
    {} & {\textcolor[rgb]{ 0,  0,  0}{NoC@85}} & {\textcolor[rgb]{ 0,  0,  0}{NoC@90}} & {\textcolor[rgb]{ 0,  0,  0}{NoC@85}} & {\textcolor[rgb]{ 0,  0,  0}{NoC@90}} \\
    \midrule
    {$0.7$} & $1.74$  & $2.02$  & $3.58$  & $5.00$ \\
    {$0.8$} & $\textbf{1.44}$  & $\textbf{1.72}$  & $\textbf{3.26}$  & $\textbf{4.82}$ \\
    {$0.9$} & $1.55$  & $1.90$  & $3.49$  & $4.97$ \\
    \bottomrule
  \end{tabular}%
  }
  \label{MaskMatchThreshold}%
\end{table}%

\begin{table}[t]
  \centering
  \caption{Averaged NoC obtained with different choices of guidance map.}
  \fontsize{9}{10}\selectfont
  \setlength{\tabcolsep}{3pt}
    \begin{tabular}{l|c|c|c|c}
    \toprule
    {\multirow{2}[0]{*}{\textbf{Variants}}} & \multicolumn{2}{c|}{Berkeley} & \multicolumn{2}{c}{DAVIS} \\
    \cline{2-5}
    {}& {\textcolor[rgb]{ 0,  0,  0}{NoC@85}} & {\textcolor[rgb]{ 0,  0,  0}{NoC@90}} & {\textcolor[rgb]{ 0,  0,  0}{NoC@85}} & {\textcolor[rgb]{ 0,  0,  0}{NoC@90}} \\
    \midrule
    {Prev. Mask}        & $\textbf{1.38}$ & $1.75$ & $3.32$ & $4.88$ \\
    {Points}            & $1.46$ & $1.81$ & $3.34$ & $4.91$ \\
    {Prev. Mask \& Points} & $1.44$ & $\textbf{1.72}$ & $\textbf{3.26}$ & $\textbf{4.82}$ \\
    \bottomrule
    \end{tabular}%
  \label{saliencyMap}%
\end{table}%

\subsection{Ablation Study}
We conduct ablation studies on the Berkeley~\cite{937655} and DAVIS~\cite{Perazzi_CVPR_2016} datasets, employing segformerB3 as the backbone for the segmentation model.

\noindent \textbf{Core Component Analysis.} Table~\ref{module} delineates the outcomes from diverse configurations of our approach. The baseline configuration excludes both the MaskMatch and TAIZ modules. Introducing either the MaskMatch or TAIZ modules distinctly enhances the NoC metrics on both datasets. Integrating both modules accentuates this improvement. Specifically, the integration of MaskMatch and TAIZ reduces the NoC@85 metric by 0.07 and 0.13, respectively, on the DAVIS dataset. When employed both modules, the decrease in NoC@85 is 0.22 on DAVIS, in comparison to employing only MaskMatch with standard bilinear interpolation for image downsampling. This substantiates the supplementary benefit of TAIZ over standard bilinear interpolation.
Fig.~\ref{figmaskzoom} illustrates the segmentation outcomes from two method variations: one incorporating TAIZ (denoted as w/ TAIZ) and the other excluding it (denoted as w/o TAIZ). Visual assessments indicate that TAIZ brings a more comprehensive target extraction.

Moreover, we try to substitute the matching regularization term in Eq.~(\ref{eq:loss-cr}) with the supervised learning term from Eq.~(\ref{eq:loss-sup}), denoted by ``Replace $L_{mr}$ with $L_{sup}$''. This alteration underperforms compared to our finalized model, underscoring the efficacy of MaskMatch.


\noindent \textbf{Choice of Threshold Value for Activating MaskMatch.}
Table~\ref{MaskMatchThreshold} analyzes the threshold, $\alpha$, for MaskMatch activation. We find 0.8 as the optimal value. Lower thresholds may produce masks far from the GT, while higher ones may limit training samples in the MaskMatch process.

\noindent \textbf{Comparison of Different Guidance Strategies in TAIZ.}
Table~\ref{saliencyMap} assesses three guidance map strategies for the TAIZ module: 1) using the click map, 2) using the prior mask, and 3) combining both. The third strategy proves most effective, capturing the target's comprehensive representation and local click details.